\documentclass[letterpaper,10pt,times,mathptm,psfig,conference]{IEEEtran}
\IEEEoverridecommandlockouts

\usepackage{cite}
\usepackage{array}
\usepackage{amsmath,amssymb,amsfonts}
\usepackage{algorithmic}
\usepackage[ruled,linesnumbered]{algorithm2e}
\usepackage[normalem]{ulem}
\usepackage{graphicx}
\usepackage{textcomp}
\usepackage{xcolor}
\usepackage{xspace}
\usepackage{multirow}
\usepackage{booktabs}
\usepackage{hyperref}
\usepackage{epsfig, subfigure, wrapfig}
\def\BibTeX{{\rm B\kern-.05em{\sc i\kern-.025em b}\kern-.08em
    T\kern-.1667em\lower.7ex\hbox{E}\kern-.125emX}}

\newcommand{\khalil}[1]{}

\author{
Md Khalid Syfullah and Alvi Ataur Khalil\\
Computer Science, Southern Illinois University Carbondale, USA\\
\{mdkhalid.syfullah, a.khalil\}@siu.edu
\vspace{-15pt}
}

\begin{document}
\title{Unit-Aware Differentially Private Quantum Machine Learning for Clinical and Wearable Health Data
\vspace{-10pt}}

\maketitle
\bstctlcite{IEEEexample:BSTcontrol}
\thispagestyle{plain}
\pagestyle{plain}

\begin{abstract}
Clinical registries and wearable sensors support early health-risk assessment, yet models trained on these records may leak sensitive information when released. Differential privacy (DP) bounds this leakage by injecting noise whose total magnitude grows with the number of trainable parameters, a cost especially severe for small health cohorts. This motivates parameter efficient models including quantum machine learning (QML), whose variational circuits can represent complex decision boundaries with few trainable parameters. It raises a question: at matched input dimensionality and privacy budget, does a quantum model retain more utility than a classical one? We investigate this question on three clinical cohorts and one wearable cohort, treating each clinical record or participant as a privacy unit. We compare capacity-matched quantum, multilayer perceptron, and logistic regression models across five privacy budgets and five seeds. The quantum model leads on three clinical cohorts by up to 0.104 ROC AUC averaged across budgets; however, trails on the wearable cohort, where the small number of participants requires substantial noise. No difference survives Holm correction across 24 tests, so these results indicate a trend rather than a quantum advantage. An analytic gradient bound shows that lower clipping thresholds can improve utility by reducing noise despite clipping bias. Finally, an experiment on IBM quantum computer (\texttt{ibm\_fez}, a 156-qubit superconducting processor) using eight fixed test cases yields mean absolute probability errors of 0.046 (private) and 0.063 (non-private) relative to simulation, supporting the feasibility of inference on noisy intermediate-scale quantum (NISQ) hardware. The code is available publicly at \url{https://github.com/TITANS-Lab-Artifacts/DP-QML-Health}.
\end{abstract}


\vspace{-5pt}
\section{Introduction}
\label{sec:introduction}
Healthcare data collection now extends well beyond hospital walls. Clinical registries record diagnoses, laboratory values, and treatment outcomes, while wearable devices continuously capture vital signs and physiological signals, including heart rate, respiration, skin conductance, temperature, and motion. Together they support automated risk stratification and continuous monitoring, improving early screening~\cite{esteva2019guide}. Machine learning (ML) enables predictive models to be trained on these data; however, the resulting models introduce a privacy risk: released model parameters can reveal whether an individual contributed data and can retain patient-specific patterns~\cite{shokri2017membership}. A trustworthy healthcare model must therefore protect patient privacy while maintaining useful predictive accuracy.

Differential privacy (DP) addresses this risk by bounding how much the output distribution of a training algorithm can change between neighboring datasets, which differ only in the data contributed by a single privacy unit~\cite{dwork2006calibrating}. The privacy unit identifies whose data the guarantee protects: one patient record in the clinical datasets and one participant in the wearable dataset. Differentially private stochastic gradient descent (DP-SGD) enforces the guarantee by clipping per-unit gradients, adding Gaussian noise, and accounting for privacy loss~\cite{mironov2017renyi}. Since DP is closed under post-processing, its guarantee remains valid when trained weights are analyzed, including by future attacks. This makes DP-SGD a suitable mechanism for protecting training data when models are released~\cite{dwork2014algorithmic}. The protection is not free, and its cost is general rather than specific to healthcare: the excess risk of DP-SGD can grow with the number of trainable parameters and decrease with the number of privacy units~\cite{bassily2014private}. Small health cohorts make this tradeoff difficult: a few hundred records or a few dozen participants may need substantial noise to meet a fixed privacy budget, while heavy clipping can distort useful gradient information.  Wearables compound this: each participant contributes many correlated windows, so treating windows as separate units inflates the unit count and overstates participant-level protection. A defensible analysis protects each participant as one unit and accepts the resulting scarcity of units~\cite{mcmahan2018learning}.



These privacy constraints motivate exploring models that can work with fewer trainable parameters. Parameterized quantum circuits are promising in this regard, as they can represent complex decision boundaries with relatively few parameters~\cite{benedetti2019parameterized}. A quantum neural network (QNN) built from such a circuit provides bounded measurement outputs, while the parameter-shift rule enables analytic gradient evaluation~\cite{schuld2019gradients}. Still, it remains unclear whether quantum machine learning (QML) can maintain predictive performance under privacy constraints. Hardware testing is also needed as gate errors, readout errors, and finite-shot uncertainty can cause predictions to differ from ideal simulation~\cite{cerezo2021variational}.

In this study, we investigate whether QNNs make more accurate predictions than classical models under DP. We define a unit-aware DP-SGD framework that protects each clinical record or all windows from one wearable participant as a single privacy unit. We evaluate the framework on three clinical datasets and one wearable dataset, comparing a QNN with a capacity-matched multilayer perceptron (MLP) and logistic regression (LR). Each model is evaluated across clipping thresholds and privacy budgets under different random seeds. We use Wilcoxon signed-rank tests~\cite{wilcoxon1945individual} with Holm correction~\cite{holm1979multiple} to assess whether observed performance differences are statistically significant to establish a potential quantum benefit in this domain. Our contributions are threefold:

\begin{itemize}
    \item We derive a QNN gradient bound to identify a threshold that avoids clipping and test whether lower thresholds improve predictive performance by reducing DP noise.
    \item We conduct an ablation study to assess how encoding, parameter count, and circuit depth affect QNN performance under DP and identify best settings on each dataset.
    \item We run selected trained QNNs on noisy intermediate-scale quantum (NISQ) hardware, \texttt{ibm\_fez} to compare their predictions with exact statevector simulation.
\end{itemize}
 

The study addresses three research questions:

\begin{itemize}
    \item \textbf{RQ1}: Does a capacity-matched QNN retain more utility than classical baselines under record-level and participant-level DP in this domain?
    \item \textbf{RQ2}: How do the clipping threshold, encoding gate, parameter count, and circuit depth affect predictive performance under DP?
    \item \textbf{RQ3}: Do predictions from NISQ hardware agree with exact statevector inference for the same trained circuits?
\end{itemize}


Paper Organization: Section~\ref{sec:background} introduces the necessary background; Section~\ref{sec:related} reviews related work; Section~\ref{sec:method} presents the methodology; Section~\ref{sec:results} reports the results; and Section~\ref{sec:conclusion} concludes the paper.

 

\section{Background}
\label{sec:background}

This section introduces the concepts underlying our work.

\subsection{Quantum Neural Networks}

A QNN embeds a parameterized quantum circuit within a hybrid classical-quantum model. Measurement yields a value mapped to a prediction, and a classical optimizer updates the gate parameters from the training loss~\cite{mitarai2018quantum}. Deep circuits complicate optimization~\cite{mcclean2018barren}, and hardware compilation expands them into noisy native operations.

\subsection{Differential Privacy}


DP guarantees that no single unit substantially affects the released model~\cite{dwork2006calibrating}. DP-SGD enforces this by clipping each unit’s contribution to an update and adding calibrated noise~\cite{abadi2016deep}, with a R{‘e}nyi accountant tracking the cumulative privacy budget~\cite{mironov2017renyi}.

\subsection{The Choice of Privacy Unit}

\begin{figure}[t]
    \centering
    \includegraphics[width=0.95\columnwidth]{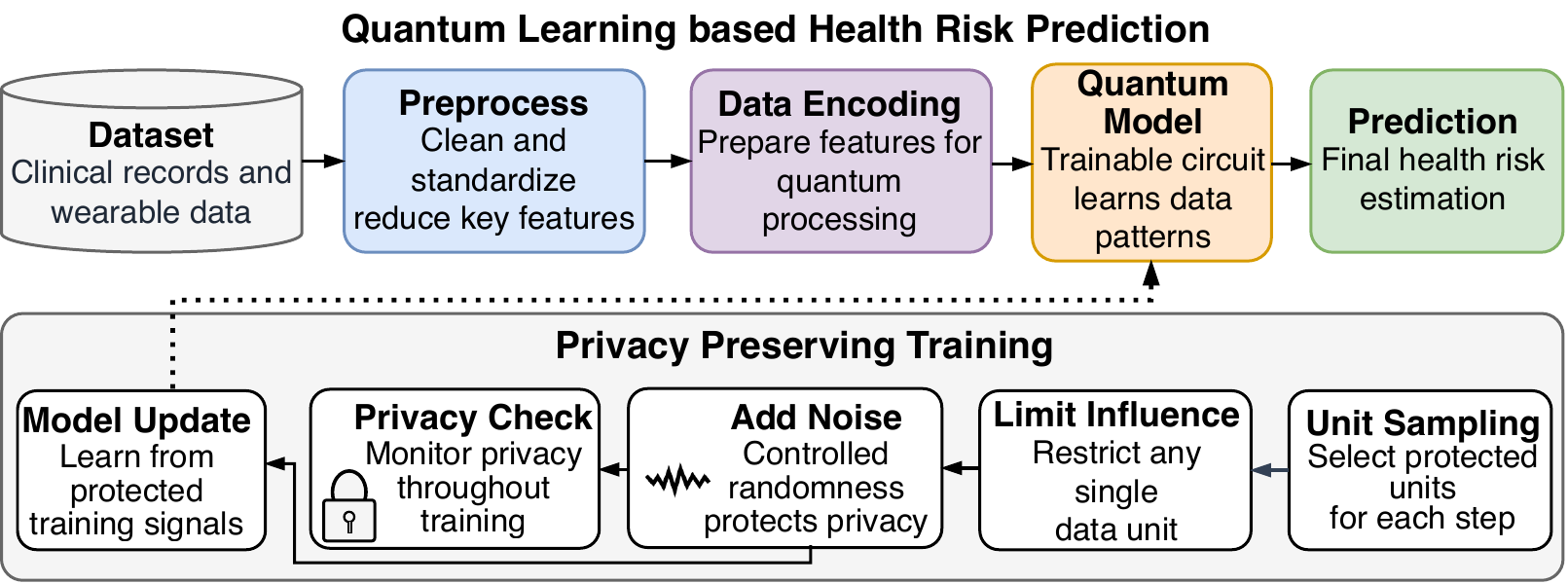}
    \vspace{-8pt}
    \caption{Quantum risk prediction with privacy-preserving training.}
    \vspace{-10pt}
    \label{fig:background}
\end{figure}

DP's protection depends on how a data unit is defined. Fig.~\ref{fig:background} shows this choice at the unit-sampling stage, prior to clipping and noise addition. A clinical row typically represents one patient encounter, and thus one protected unit. In wearable data, however, one participant contributes many correlated windows. Protecting windows individually still leaks participant-specific physiology, device placement, and labeling patterns across a participant's other windows. We therefore group the 438 windows in WESAD dataset~\cite{schmidt2018wesad} by participant into 15 privacy units, following user-level private learning~\cite{mcmahan2018learning}, which requires more noise at any fixed privacy budget.


\section{Literature Review}
\label{sec:related}

Privacy-preserving quantum learning has adopted several DP strategies: Gaussian-noise clipping in variational classifiers~\cite{watkins2023quantum}, private teacher aggregation with a student trained on noisy votes~\cite{watkins2024qpate}, federated learning that protects clients without exchanging raw records~\cite{rofougaran2024federated}, DP-SGD on variational circuits for time-series forecasting~\cite{chen2025qdpts}, and analyses of clipping and calibrated noise in hybrid pipelines~\cite{sedrakyan2026private}. These studies demonstrate DP training of variational circuits, mainly on image, synthetic, or non-clinical benchmarks.
Quantum gradients can disclose input structure under some circuit conditions: bounded outputs alone do not guarantee privacy, and training requires an explicit mechanism and accountant~\cite{heredge2025characterizing}. The parameter-shift rule can tighten sensitivity bounds and reduce the noise required~\cite{ngo2026qshiftdp}. These two studies examine gradient leakage and its mitigation yet didn't adapt an analogous gradient bound to identify a clipping threshold that preserves all per-unit gradients.

Quantum kernels and variational classifiers have reached several clinical tasks, including classification on NISQ hardware~\cite{moradi2022clinical}, ischemic heart-disease prediction from electronic records~\cite{maheshwari2023quantum}, and heart-disease prediction benchmarked against classical baselines~\cite{babu2024heart}. A systematic review reports methodological weaknesses, limited hardware validation, and insufficient evidence of quantum advantage~\cite{gupta2024systematic}. These results demonstrate the feasibility of quantum models for clinical tasks, yet not establish hardware-validated quantum advantage.

A remaining gap in QML is the joint evaluation of a formal privacy mechanism with reported privacy expenditure, comparisons with capacity-matched classical baselines, and record- and participant-level protection of health data. To the best of our knowledge, no prior study includes all three and executes a model trained with DP on real quantum hardware. We address this gap by using capacity-matched models, keeping results across privacy budgets grouped by random seed so that evaluations on the same data split are not treated as independent, correcting statistical tests for multiple comparisons, and comparing quantum hardware predictions with exact statevector simulation.

\section{Methodology}
\label{sec:method}

This section describes the models, training process, datasets, and evaluation procedure. Table~\ref{tab:notation} summarizes the notation used throughout the paper.

\begin{table}[hbt!]
\centering
\caption{Summary of notation.}
\vspace{-5pt}
\setlength{\tabcolsep}{3pt}
\renewcommand{\arraystretch}{1.05}
\scriptsize
\begin{tabular}{|c|l|c|l|}
\hline
\textbf{Sym.} & \textbf{Meaning} & \textbf{Sym.} & \textbf{Meaning} \\
\hline
$u$ & Privacy-unit index & $d$ & Qubits, components \\
\hline
$n$ & Number of private units & $L$ & Variational layers \\
\hline
$q$ & Poisson sampling rate & $\mathbf{x},y$ & Input feature, binary label \\
\hline
$\mathcal{B}_t$ & Units sampled at $t$ & $p_{\boldsymbol{\theta}}$ & Positive-class prob. \\
\hline
$\varepsilon,\delta$ & Budget, failure prob. & $\ell_u$ & Loss of unit $u$ \\
\hline
$\sigma$ & Noise multiplier & $\mathbf{g}_u,\bar{\mathbf{g}}_u$ & Raw, clipped gradient \\
\hline
$C$ & Clipping threshold & $\boldsymbol{\xi}_t$ & Gaussian noise at $t$ \\
\hline
$\boldsymbol{\theta}$ & Trainable parameters & $\eta$ & Learning rate \\
\hline
$P$ & Number of parameters & $Z_0$ & Pauli $Z$, first qubit \\
\hline
\end{tabular}
\vspace{-10pt}
\label{tab:notation}
\end{table}

\subsection{Models Used in This Study}

\begin{figure}[t]
    \centering
    \includegraphics[width=0.95\columnwidth]{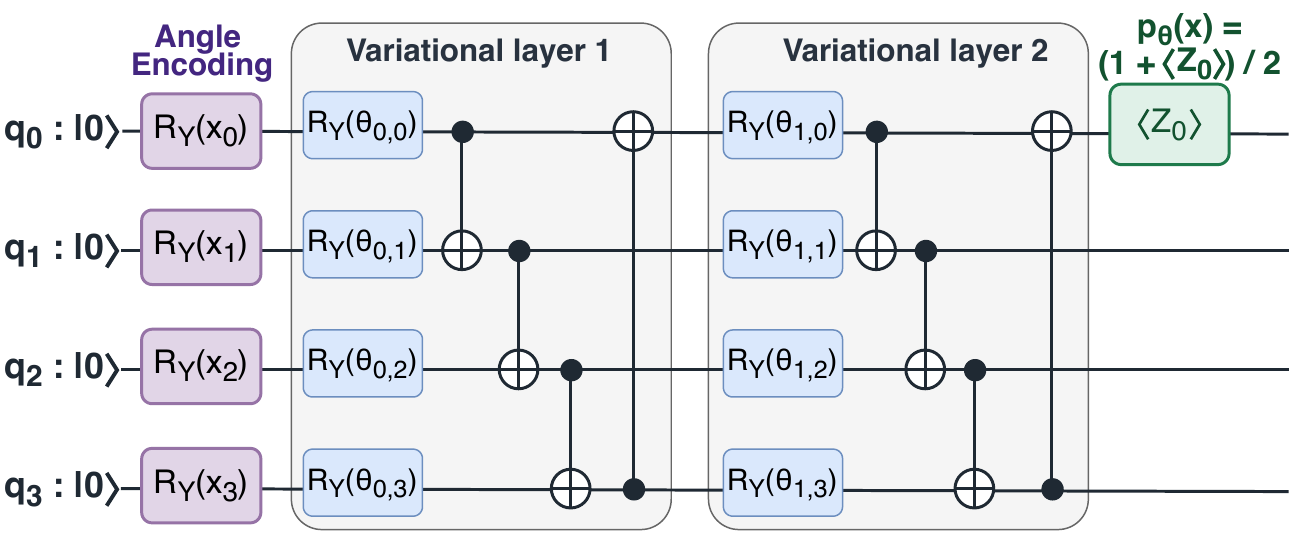}
    \vspace{-10pt}
    \caption{Architecture of the variational quantum classifier.}
    \vspace{-15pt}
    \label{fig:circuit}
\end{figure}

Our circuit, shown in Fig.~\ref{fig:circuit}, uses $d=4$ qubits to encode the four input components and $L=2$ trainable layers with $P=8$ angles. Each qubit starts in $\lvert 0\rangle$ and encodes one transformed feature through an $R_Y$ rotation. Each layer then applies four trainable $R_Y$ rotations and a ring of controlled-NOT (CNOT) gates connecting qubits in the order $0\rightarrow1\rightarrow2\rightarrow3\rightarrow0$. We obtain the positive-class probability $p_{\boldsymbol{\theta}}$ by rescaling the expectation of $Z_0$, the Pauli $Z$ operator on the first qubit as:

\vspace{-6pt}
\begin{equation}
p_{\boldsymbol{\theta}}(\mathbf{x})
=\frac{1+\left\langle\psi_{\boldsymbol{\theta}}(\mathbf{x})\right|
Z_0\left|\psi_{\boldsymbol{\theta}}(\mathbf{x})\right\rangle}{2},
\label{eq:qprob}
\end{equation}

\noindent where $\left|\psi_{\boldsymbol{\theta}}(\mathbf{x})\right\rangle$ is the quantum state produced by the circuit for input features $\mathbf{x}$ and trainable angles $\boldsymbol{\theta}$. For classical models, we use a capacity-matched MLP and LR. Capacity matching means using the same input dimensionality: each model receives the same $d=4$ principal component analysis (PCA) components. The MLP has four inputs, two hyperbolic-tangent hidden units, one sigmoid output, totaling $4\times2+2+2\times1+1=13$ parameters. The differentiable LR model has $P=5$ parameters and provides a linear baseline. We use ROC AUC as the primary metric, with PR-AUC, accuracy, mean squared error (MSE), and Brier score as additional metrics.

\subsection{System Overview}

\begin{figure*}[!t]
    \centering
    \includegraphics[width=0.75\textwidth]{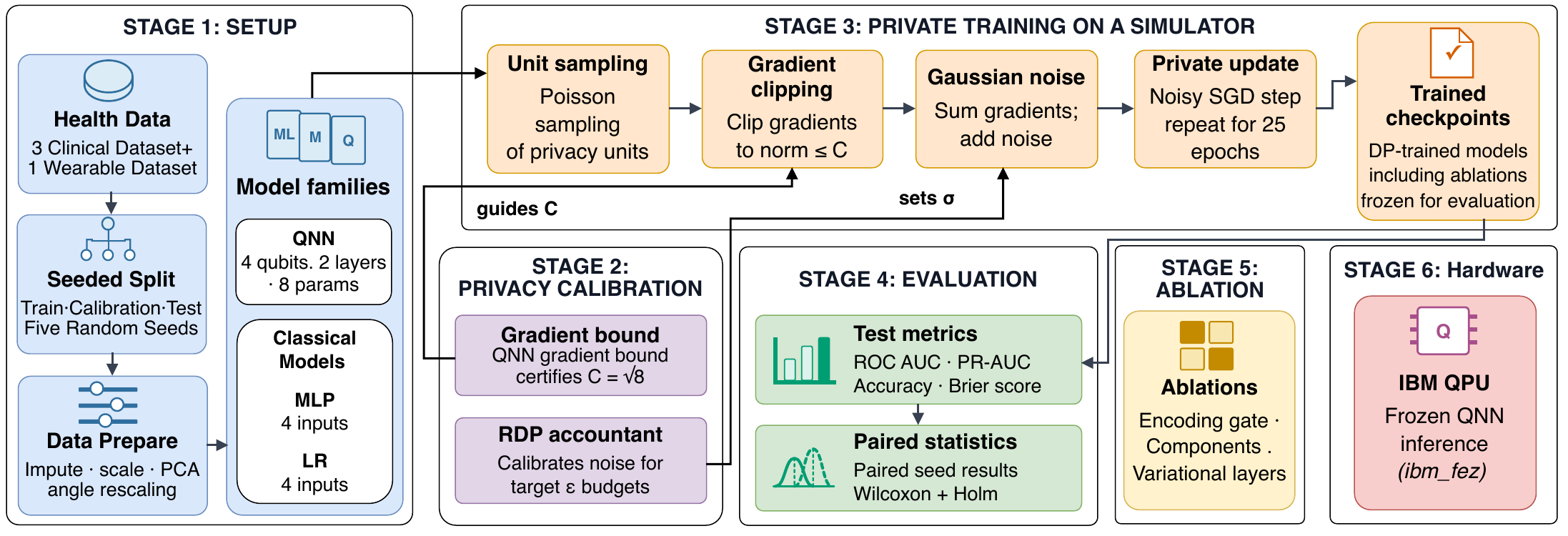}
    \vspace{-10pt}
    \caption{Overview of the unit-aware DP-SGD framework, including setup, training, statistical evaluation, circuit ablations, and quantum hardware inference.}
    \vspace{-15pt}
    \label{fig:system}
\end{figure*}

Fig.~\ref{fig:system} presentes of the proposed system, which contains six different stages. First, the setup stage splits each dataset into training, calibration, and test sets under repeated random seeds, performs preprocessing, and prepares the features and models. The privacy calibration stage fixes the gradient clipping norm and the noise level for each privacy budget. Then, the private training stage trains every model with DP-SGD on a simulator through multiple substages and saves the checkpoints. The evaluation stage is for performance evaluation and compares the trained models with paired statistical tests. Finally, the ablation stage isolates the effect of each QNN design choice, and the hardware stage runs the trained QNN on an IBM quantum processor; both serve as additional experiments.

\subsection{Unit-Aware Private Optimization}

We apply the same unit-aware DP-SGD procedure to the QNN, MLP, and LR models. For clinical data, the loss of one record is $\ell_u=(p_{\boldsymbol{\theta}}(\mathbf{x}_u)-y_u)^2$. For WESAD, $\ell_u$ is the mean loss across all windows from one participant. At each training step, we independently sample each unit with probability $q$. We then clip its gradient so that its Euclidean norm does not exceed the threshold $C$:

\vspace{-6pt}
\begin{equation}
\bar{\mathbf{g}}_u
=\mathbf{g}_u\,
\min\!\left(1,\frac{C}{\lVert\mathbf{g}_u\rVert_2}\right),
\qquad
\mathbf{g}_u=\nabla_{\boldsymbol{\theta}}\ell_u .
\label{eq:clip}
\end{equation}

For the sampled units $\mathcal{B}_t$, we update the parameters using clipped gradients and Gaussian noise, normalized by the expected batch size $nq$:
\vspace{-6pt}
\begin{equation}
\boldsymbol{\theta}_{t+1}
=\boldsymbol{\theta}_t-\frac{\eta}{nq}
\left(
\sum_{u\in\mathcal{B}_t}\bar{\mathbf{g}}_u
+\boldsymbol{\xi}_t
\right).
\label{eq:dpsgd}
\end{equation}
Here, $\eta$ is the learning rate and $\boldsymbol{\xi}_t\sim\mathcal{N}(\mathbf{0},\sigma^2C^2\mathbf{I})$ is Gaussian noise with standard deviation $\sigma C$ per coordinate, where $\sigma$ is the noise multiplier and $\mathbf{I}$ is the identity matrix. Opacus selects $\sigma$ for the target privacy budget using R{\'e}nyi accounting for the sampled Gaussian mechanism~\cite{yousefpour2021opacus}. 
Every run uses $\delta=10^{-5}$, 25 epochs, and an expected batch size capped at 32 units.

\subsection{Gradient Sensitivity Certificate}

Let $z_{\boldsymbol{\theta}}=\langle Z_0\rangle\in[-1,1]$. The parameter-shift rule bounds the magnitude of its derivative with respect to each angle by one. Since $p_{\boldsymbol{\theta}},y\in[0,1]$, each coordinate $\partial\ell_u/\partial\theta_k$ of the single-record MSE gradient satisfies:

\vspace{-6pt}
\begin{equation}
\left|\frac{\partial\ell_u}{\partial\theta_k}\right|
=\left|(p_{\boldsymbol{\theta}}-y)\frac{\partial z_{\boldsymbol{\theta}}}{\partial\theta_k}\right|\leq 1,
\qquad
\lVert\nabla_{\boldsymbol{\theta}}\ell_u\rVert_2\leq\sqrt{P}.
\label{eq:certificate}
\end{equation}

The norm bound applies to a participant's mean loss since its gradient averages window-level gradients. For the primary QNN, $P=8$ gives the certified threshold $C=\sqrt{8}$. Clipping is inactive at or above this threshold. For fixed $\sigma$, a lower threshold reduces the noise scale $\sigma C$ in Eq.~\eqref{eq:dpsgd}; however, it can introduce bias by shrinking gradients whose norms exceed $C$ in Eq.~\eqref{eq:clip}.

\subsection{Datasets and Calibration}
We divide each clinical dataset into stratified 60/20/20 training, calibration, and test subsets as shown on Table~\ref{tab:data}. For each seed, WESAD's 15 wrist-sensor participants are split into nine for private training, three for calibration, and three for testing, with 438 one-minute windows in total. We use median imputation to fill missing values and standardization to put features on a common scale. PCA reduces the input to four components for the four-qubit circuit, and $[0,\pi]$ rescaling maps these components to encoding angles.

\subsection{Evaluation Protocol}

We evaluate models on four datasets with five random seeds to assess variation across data splits and training randomness. Each seed defines shared training, calibration, and test splits for paired model comparisons. QNN simulation uses PennyLane \texttt{lightning.qubit} with adjoint differentiation. Table~\ref{tab:evaluation_setup} summarizes the setup. Three nonprivate QNN runs repeat one configuration, providing a baseline for each clipping-threshold column. WESAD has seven input features, so eight-component configurations are excluded. We average results across five budgets within each seed.

\begin{table}[t]
\centering
\caption{Datasets and privacy settings at target $\varepsilon=1$.}
\vspace{-5pt}
\label{tab:data}
\setlength{\tabcolsep}{2pt}
\renewcommand{\arraystretch}{1}
\scriptsize
\begin{tabular}{@{}lrrrlrrr@{}}
\toprule
Dataset & Samp. & In. & Pos.\% & Unit & Units(train/cal./test) & $q$ & $\sigma$ \\
\midrule
Pima Diabetes~\cite{smith1988pima} & 768 & 8 & 34.9 & record & 460/154/154 & 0.067 & 5.39 \\
UCI Heart~\cite{detrano1989heart} & 303 & 13 & 45.9 & record & 181/61/61 & 0.167 & 8.44 \\
Heart Failure~\cite{chicco2020heartfailure} & 299 & 12 & 32.1 & record & 179/60/60 & 0.167 & 8.44 \\
WESAD~\cite{schmidt2018wesad} & 438 & 7 & 35.4 & particip. & 9/3/3 & 1.000 & 20.31 \\
\bottomrule
\end{tabular}
\vspace{-10pt}
\end{table}

\begin{table}[!t]
\centering
\caption{Training and statistical evaluation setup.}
\vspace{-5pt}
\label{tab:evaluation_setup}
\setlength{\tabcolsep}{2pt}
\renewcommand{\arraystretch}{1.08}
\scriptsize
\begin{tabular}{
|>{\centering\arraybackslash}m{0.18\columnwidth}
|>{\centering\arraybackslash}m{0.40\columnwidth}
|>{\centering\arraybackslash}m{\dimexpr0.42\columnwidth-6\tabcolsep-4\arrayrulewidth\relax}|}
\hline
\textbf{Category} & \textbf{Settings} & \textbf{Configurations} \\
\hline
\textbf{Private training}
&
Five privacy budgets:
$\varepsilon\in\{0.5,1,2,4,8\}$.
\newline
Six model--clipping combinations:
QNN $C\in\{0.5,1,\sqrt{8}\}$;
MLP $C\in\{0.5,1\}$;
LR $C=1$.
&
6 model--clipping combinations $\times$ 5 budgets $\times$ 4 datasets $\times$ 5 seeds.
\newline
\textbf{Total: 600 runs}
\\
\hline
\textbf{Nonprivate training}
&
Three QNN runs for the columns labeled $C=0.5$, $1$, and $\sqrt{8}$; one MLP and one LR.
&
5 runs $\times$ 4 datasets $\times$ 5 seeds.
\newline
\textbf{Total: 100 runs}
\\
\hline
\textbf{Circuit ablations}
&
Privacy budget: $\varepsilon=1$;
Clipping threshold: $C=1$;
Encoding gates: $R_X$, $R_Y$;
PCA components: 4, 6, 8;
Trainable layers: 1, 2, 3;
\newline
\textbf{Circuit Settings=2x3x3 =18}
&
18 circuit settings $\times$ 4 datasets $\times$ 5 seeds (excluding 30 WESAD runs requiring 8 components).
\newline
\textbf{Total: 330 runs}
\\
\hline
\textbf{Statistical evaluation}
&
QNN--MLP results at $C=1$, paired by seed.
Two-sided Wilcoxon signed-rank tests.
Descriptive 95\% intervals from 10,000 paired bootstrap resamples.
&
20 per-budget comparisons and 4 comparisons across budgets.
\newline
\textbf{Total: 24 tests}, with Holm correction~\cite{holm1979multiple}.
\\
\hline
\multicolumn{3}{|c|}{\textbf{Total training runs:} $600+100+330=1{,}030$} \\
\hline
\end{tabular}
\vspace{-10pt}
\end{table}


\section{Experimental Results}
\label{sec:results}

This section presents the experimental setup, model performance under DP, statistical comparisons, circuit design results, and quantum hardware validation.

\subsection{Experimental Setup}

Simulations used PennyLane 0.38.0, PyTorch 2.8.0, Opacus 1.6.0, scikit-learn 1.6.1, and Python 3.9. Quantum hardware execution used Qiskit 2.4.2 and IBM Runtime 0.46.1. Results are reported as means, with sample standard deviations in parentheses. Actual privacy budgets differ from their targets by at most 0.0082. Classical computations ran on a workstation with an AMD Ryzen Threadripper PRO 7955WX CPU (16 cores, 32 threads, up to 5.4~GHz), two NVIDIA RTX PRO 6000 Blackwell GPUs with 192~GB of combined VRAM, and 32~GB of system RAM.

\subsection{Predictive Performance Across Privacy Budgets}

\begin{figure}[t]
    \centering
    \includegraphics[width=0.8\columnwidth]{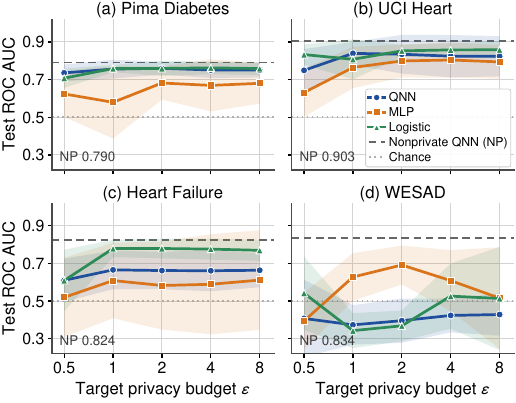}
    \vspace{-10pt}
    \caption{ROC AUC performance across privacy budgets at $C=1$. Bands show one standard deviation; dashed and dotted lines indicate the nonprivate QNN and random baseline, respectively.}
    \label{fig:privacy}
    \vspace{-16pt}
\end{figure}

Table~\ref{tab:master} reports results for all model-clipping combinations and privacy budgets. Fig.~\ref{fig:privacy} shows the primary comparison at $C=1$. Without DP, QNN and MLP differ by less than 0.010 AUC on the three clinical datasets. On WESAD, the QNN leads by 0.112 AUC with less variation across seeds.

\begin{table*}[!t]
\centering
\caption{Private utility across models, clipping thresholds, and privacy budgets: Private test ROC AUC (standard deviation) versus target budget $\varepsilon$; column labels give the clipping threshold $C$.}

\vspace{-10pt}
\label{tab:master}
\setlength{\tabcolsep}{2.5pt}
\renewcommand{\arraystretch}{0.95}
\scriptsize

\begin{minipage}[t]{0.485\textwidth}
\centering
\textbf{(a) Pima Diabetes}
\resizebox{\linewidth}{!}{%
\begin{tabular}{@{}lcccccc@{}}
\toprule
$\varepsilon$
& QNN $.5$
& QNN $1$
& QNN $\sqrt{8}$
& MLP $.5$
& MLP $1$
& LR $1$ \\
\midrule
$\infty$
& .790 (.017)
& .790 (.017)
& .790 (.017)
& .796 (.016)
& .796 (.016)
& .798 (.016) \\

.5
& .760 (.032)
& .735 (.044)
& .668 (.122)
& .535 (.133)
& .624 (.122)
& .707 (.051) \\

1
& .763 (.033)
& .756 (.050)
& .659 (.139)
& .573 (.111)
& .579 (.192)
& .758 (.019) \\

2
& .756 (.035)
& .760 (.034)
& .705 (.097)
& .558 (.160)
& .682 (.088)
& .759 (.040) \\

4
& .754 (.037)
& .751 (.042)
& .752 (.028)
& .562 (.140)
& .669 (.138)
& .762 (.034) \\

8
& .757 (.035)
& .750 (.039)
& .752 (.041)
& .551 (.165)
& .680 (.108)
& .759 (.034) \\
\midrule
Avg.
& \textbf{.758 (.034)}
& .750 (.040)
& .707 (.039)
& .556 (.131)
& .647 (.095)
& .749 (.033) \\
\bottomrule
\end{tabular}%
}
\end{minipage}
\hfill
\begin{minipage}[t]{0.485\textwidth}
\centering
\textbf{(b) UCI Heart}

\resizebox{\linewidth}{!}{%
\begin{tabular}{@{}lcccccc@{}}
\toprule
$\varepsilon$
& QNN $.5$
& QNN $1$
& QNN $\sqrt{8}$
& MLP $.5$
& MLP $1$
& LR $1$ \\
\midrule
$\infty$
& .903 (.033)
& .903 (.033)
& .903 (.033)
& .909 (.022)
& .909 (.022)
& .900 (.012) \\

.5
& .818 (.046)
& .749 (.113)
& .809 (.078)
& .797 (.097)
& .631 (.129)
& .832 (.053) \\

1
& .832 (.101)
& .839 (.061)
& .712 (.225)
& .711 (.138)
& .764 (.110)
& .808 (.106) \\

2
& .839 (.081)
& .834 (.103)
& .849 (.048)
& .762 (.113)
& .799 (.088)
& .852 (.033) \\

4
& .826 (.113)
& .824 (.112)
& .812 (.135)
& .786 (.111)
& .804 (.094)
& .857 (.034) \\

8
& .824 (.113)
& .824 (.107)
& .838 (.087)
& .782 (.102)
& .794 (.090)
& .858 (.047) \\
\midrule
Avg.
& .828 (.085)
& .814 (.083)
& .804 (.077)
& .767 (.105)
& .758 (.072)
& \textbf{.842 (.052)} \\
\bottomrule
\end{tabular}%
}
\end{minipage}

\vspace{3pt}
\begin{minipage}[t]{0.485\textwidth}
\centering
\textbf{(c) Heart Failure}

\resizebox{\linewidth}{!}{%
\begin{tabular}{@{}lcccccc@{}}
\toprule
$\varepsilon$
& QNN $.5$
& QNN $1$
& QNN $\sqrt{8}$
& MLP $.5$
& MLP $1$
& LR $1$ \\
\midrule
$\infty$
& .824 (.045)
& .824 (.045)
& .824 (.045)
& .816 (.059)
& .816 (.059)
& .804 (.055) \\

.5
& .648 (.091)
& .610 (.115)
& .601 (.145)
& .575 (.256)
& .520 (.210)
& .609 (.163) \\

1
& .649 (.108)
& .666 (.102)
& .701 (.161)
& .608 (.244)
& .609 (.200)
& .779 (.046) \\

2
& .656 (.114)
& .662 (.099)
& .699 (.104)
& .580 (.259)
& .583 (.234)
& .779 (.052) \\

4
& .655 (.113)
& .661 (.109)
& .691 (.109)
& .573 (.267)
& .589 (.266)
& .776 (.056) \\

8
& .651 (.105)
& .664 (.090)
& .666 (.101)
& .582 (.267)
& .612 (.264)
& .769 (.056) \\
\midrule
Avg.
& .652 (.102)
& .653 (.094)
& .672 (.095)
& .584 (.258)
& .582 (.231)
& \textbf{.743 (.059)} \\
\bottomrule
\end{tabular}%
}
\end{minipage}
\hfill
\begin{minipage}[t]{0.485\textwidth}
\centering
\textbf{(d) WESAD}

\resizebox{\linewidth}{!}{%
\begin{tabular}{@{}lcccccc@{}}
\toprule
$\varepsilon$
& QNN $.5$
& QNN $1$
& QNN $\sqrt{8}$
& MLP $.5$
& MLP $1$
& LR $1$ \\
\midrule
$\infty$
& .834 (.065)
& .834 (.065)
& .834 (.065)
& .722 (.160)
& .722 (.160)
& .729 (.173) \\

.5
& .455 (.179)
& .408 (.206)
& .534 (.312)
& .541 (.155)
& .395 (.150)
& .542 (.195) \\

1
& .541 (.202)
& .374 (.105)
& .628 (.172)
& .505 (.228)
& .628 (.126)
& .343 (.093) \\

2
& .467 (.124)
& .396 (.117)
& .397 (.178)
& .528 (.202)
& .691 (.103)
& .369 (.082) \\

4
& .449 (.124)
& .424 (.090)
& .495 (.165)
& .628 (.193)
& .609 (.162)
& .526 (.173) \\

8
& .452 (.114)
& .429 (.112)
& .472 (.096)
& .593 (.283)
& .516 (.269)
& .514 (.274) \\
\midrule
Avg.
& .473 (.120)
& .406 (.085)
& .505 (.073)
& .559 (.172)
& \textbf{.568 (.115)}
& .459 (.054) \\
\bottomrule
\end{tabular}%
}
\end{minipage}
\vspace{-10pt}
\end{table*}

\begin{table}[!t]
\centering
\caption{Paired QNN-minus-MLP ROC AUC differences at $C=1$ ($p$: uncorrected Wilcoxon; Holm-adjusted $p=1.000$ throughout).}
\vspace{-5pt}
\label{tab:stats}
\setlength{\tabcolsep}{2pt}
\renewcommand{\arraystretch}{1.0}
\scriptsize
\begin{tabular}{@{}lccccccrc@{}}
\toprule
& \multicolumn{5}{c}{Difference at budget $\varepsilon$} &
\multicolumn{3}{c}{Curve average by seed} \\
\cmidrule(lr){2-6}\cmidrule(lr){7-9}
Dataset & $0.5$ & $1$ & $2$ & $4$ & $8$ &
$\Delta$ & 95\% interval & $p$ \\
\midrule
Pima & $+.112$ & $+.176$ & $+.078$ & $+.082$ & $+.070$ & $\mathbf{+.104}$ & $[+.032,+.175]$ & .125 \\
UCI Heart & $+.118$ & $+.075$ & $+.036$ & $+.020$ & $+.030$ & $\mathbf{+.056}$ & $[+.000,+.111]$ & .313 \\
Heart Fail. & $+.090$ & $+.057$ & $+.079$ & $+.072$ & $+.052$ & $\mathbf{+.070}$ & $[-.033,+.188]$ & .813 \\
WESAD & $+.013$ & $-.254$ & $-.295$ & $-.185$ & $-.087$ & $\mathbf{-.162}$ & $[-.234,-.075]$ & .125 \\
\bottomrule
\end{tabular}
\vspace{-10pt}
\end{table}

Three patterns emerge. First, the QNN retains more performance than MLP on the clinical datasets. At $C=1$, its average AUC across budgets is within 0.040 of its nonprivate result on Pima and 0.089 on UCI Heart, compared with MLP gaps of 0.149 and 0.151. At $\varepsilon=0.5$, QNN reaches 0.735 AUC on Pima, compared with 0.624 for MLP. Its variation across seeds is also smaller: 0.040 versus 0.095 on Pima and 0.094 versus 0.231 on Heart Failure.
Second, LR has the highest average AUC across budgets on two clinical datasets. The QNN improvement over MLP therefore does not extend to every classical baseline.
Third, WESAD shows the opposite pattern. At $\varepsilon=1$, its noise multiplier is 20.31, compared with 5.39 on Pima and 8.44 on the heart datasets. Performance is unstable and often near random ranking. At $C=1$, MLP leads with an average AUC of 0.568 versus 0.406 for QNN. Neither model performs reliably in this setting.

\subsection{Statistical Evidence for Model Differences}
Table~\ref{tab:stats} reports paired QNN-minus-MLP AUC differences. The QNN leads at all 15 clinical privacy budget points and trails at four of five WESAD points. Bootstrap intervals for the average difference across budgets exclude zero on Pima and WESAD; UCI Heart's lower bound is close to zero before rounding ($+0.0005$). With five paired observations, the smallest possible exact two-sided signed-rank $p$-value is 0.0625, above the 0.05 significance threshold. Every Holm-adjusted $p$-value equals 1.000. Thus, intervals excluding zero do not establish significance under the corrected tests.
\vspace{-10pt}
\begin{figure}[hbt!]
    \centering
    \includegraphics[width=0.7\columnwidth]{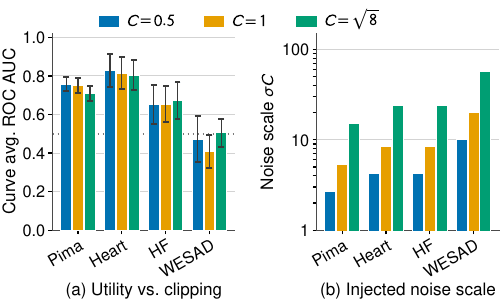}
    \vspace{-10pt}
    \caption{QNN sensitivity to the clipping threshold $C$.}
    \vspace{-12pt}
    \label{fig:clipping}
\end{figure}
\vspace{-2pt}
\subsection{Effect of the Clipping Threshold}
The certified threshold $C=\sqrt{8}$ does not give the best performance on every dataset, as shown in Fig.~\ref{fig:clipping}. It gives the highest average AUC across budgets on Heart Failure, 0.672, and reaches 0.505 on WESAD. On Pima, however, it gives 0.707 compared with 0.758 at $C=0.5$. Increasing $C$ from 0.5 to $\sqrt{8}$ increases the noise scale by 5.66 at fixed $\sigma$, showing the cost of avoiding clipping. Measurements on the calibration data support this interpretation. Mean QNN gradient norms range from 0.170 to 0.316, with a maximum observed norm of 0.685. No gradients are clipped at $C=1$. At $C=0.5$, the clipped fraction reaches 0.174 on Pima and 0.131 on Heart Failure. These measurements are taken after training on public calibration data; they do not measure clipping during private training. All observed norms remain below $\sqrt{8}$.

\subsection{Effect of Circuit Design}

\begin{figure}[hbt!]
    \centering
    \includegraphics[width=0.99\columnwidth]{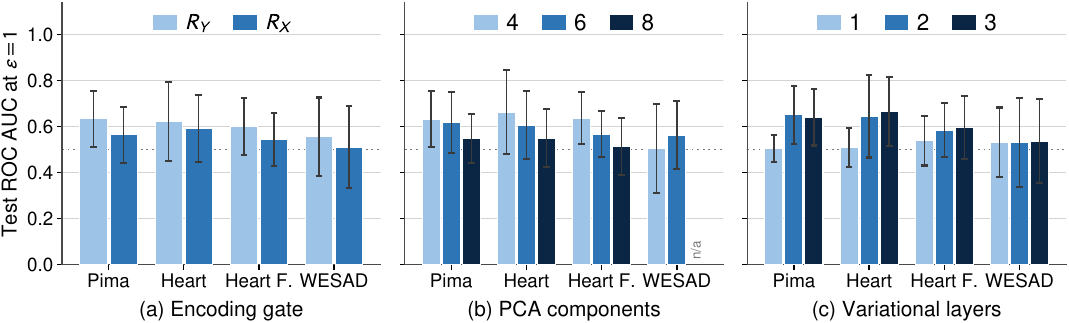}
    \vspace{-20pt}
    \caption{Private QNN circuit ablation at $\varepsilon=1$ and $C=1$.}
    \label{fig:ablation}
    \vspace{-15pt}
\end{figure}

Fig.~\ref{fig:ablation} shows the average effect of each design choice over other settings and seeds. Across the clinical datasets, $R_Y$ exceeds $R_X$ by 0.030--0.068 AUC, four components exceed eight by 0.085--0.122, and two layers exceed one by 0.045--0.146. A third layer adds little. WESAD favors six components, its only setting above the base circuit.
\vspace{-3pt}

\subsection{Prediction Agreement on IBM Quantum Hardware}
\begin{figure}[t]
    \centering
    \includegraphics[width=0.75\columnwidth]{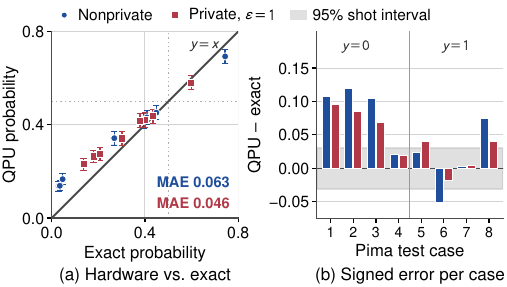}
    \vspace{-10pt}
    \caption{Hardware (QPU) versus statevector predictions.}
    \vspace{-20pt}
    \label{fig:ibm}
\end{figure}

We test the seed-zero Pima nonprivate checkpoint and the private checkpoint at $C=1$ and target $\varepsilon=1$, with actual privacy budget 0.995. We select the first four negative and first four positive test cases. A Qiskit Estimator runs both checkpoints on all eight cases in one \texttt{ibm\_fez} job with 1,024 shots, with mitigation and twirling disabled. Compilation at optimization level three increases circuit depth from 11 to 59, with 103 instructions. The job completes all 16 model-case bindings in six quantum seconds. Fig.~\ref{fig:ibm} compares hardware and exact statevector probabilities with 95\% shot intervals. Hardware predictions closely match exact inference, with MAE 0.046 for the private model and 0.063 for the nonprivate model. Both checkpoints reproduce eight simulated labels and classify five cases correctly.

\subsection{Discussion and Limitations}
For RQ1, the QNN exceeds the capacity-matched MLP on clinical datasets by 0.104, 0.056, and 0.070 average AUC across budgets; however, it trails by 0.162 on WESAD. No corrected test is significant, and LR remains competitive; therefore, results remain descriptive. For RQ2, performance depends on privacy-unit count, clipping threshold, and circuit design. The certified threshold avoids clipping while increasing noise. Shallow four-component circuits perform well on clinical datasets, while WESAD benefits from six components. For RQ3, hardware predictions closely match simulation for both checkpoints. These findings have two limitations. First, WESAD is small, which may contribute to the poor performance. A larger wearable dataset is therefore needed to assess real performance. Second, our hardware evaluation demonstrates agreement between hardware and simulated predictions for only one dataset and eight cases. However, it does not establish whether this agreement holds across other settings.

\section{Conclusion}
\label{sec:conclusion}
We presented a benchmark for differentially private quantum learning on clinical and wearable health data. The framework protects each clinical record or wearable participant and compares models at matched input dimensionality. The QNN exceeded the matched MLP on clinical datasets by up to 0.104 average ROC AUC across privacy budgets; however, it performed worse on wearable data, while LR remained competitive. No model difference was significant after Holm correction. The gradient bound showed that avoiding clipping can increase noise, and circuit ablations showed that design choices affect performance under DP. Hardware predictions closely matched exact simulation, with a mean absolute probability error of 0.046 for the private checkpoint. These findings support feasible inference on quantum hardware; however, they do not establish quantum advantage. Future work should evaluate larger cohorts under preregistered analyses, investigate alternative private training mechanisms, and tighten sensitivity bounds. Hardware evaluations should be repeated across devices and sessions with recorded physical-qubit layouts.

\bibliographystyle{IEEEtran}
\bibliography{References}

\end{document}